\PassOptionsToPackage{table}{xcolor}
\documentclass[runningheads]{llncs}

\usepackage{eccv}

\usepackage{eccvabbrv}

\usepackage[table]{xcolor}

\usepackage{graphicx}
\usepackage{booktabs}
\usepackage{physics}
\usepackage{multirow}
\usepackage{pgfplots}
\usepackage{pgfplotstable}
\usepackage{filecontents}
\usepackage{tabularx}
\usepackage{amssymb}
\usepackage{pifont}
\usepackage{hyphenat}

\usepackage[accsupp]{axessibility}  

\newcommand{\cmark}{\ding{51}}%
\newcommand{\xmark}{{\color{lightgray}\ding{55}}}

\newcommand{\mmpose}{MMpose }

\newcolumntype{P}[1]{>{\centering\arraybackslash}p{#1}}  

\usepackage[pagebackref,breaklinks,colorlinks,citecolor=eccvblue]{hyperref}

\usepackage{orcidlink}

\begin{document}

\title{MPL: Lifting 3D Human Pose from Multi-view 2D Poses} 


\author{Seyed Abolfazl Ghasemzadeh\inst{1}\orcidlink{0000-0001-7111-5778} \and
Alexandre Alahi\inst{2}\orcidlink{0000-0002-5004-1498} \and
Christophe De Vleeschouwer\inst{1}\orcidlink{0000-0001-5049-2929}}

\authorrunning{S.A.~Ghasemzadeh et al.}

\institute{Université Catholique de Louvain (UCLouvain), ICTEAM/ELEN, Belgium 
\and
Ecole Polytechnique Fédérale de Lausanne (EPFL), Switzerland
}

\maketitle

\begin{abstract}
  Estimating 3D human poses from 2D images is challenging due to occlusions and projective acquisition. Learning-based approaches have been largely studied to address this challenge, both in single and multi-view setups. These solutions however fail to generalize to real-world cases due to the lack of (multi-view) 'in-the-wild' images paired with 3D poses for training. For this reason, we propose combining 2D pose estimation, for which large and rich training datasets exist, and 2D-to-3D pose lifting, using a transformer-based network that can be trained from synthetic 2D-3D pose pairs. Our experiments demonstrate decreases up to $45\%$ in MPJPE errors compared to the 3D pose obtained by triangulating the 2D poses. The framework's source code is available at \url{https://github.com/aghasemzadeh/OpenMPL}.
\end{abstract}

\section{Introduction}
\label{sec:intro}

\begin{figure*}[t]
    \centering
    \includegraphics[width=\textwidth,trim={1cm 9cm 0cm 10cm},clip]{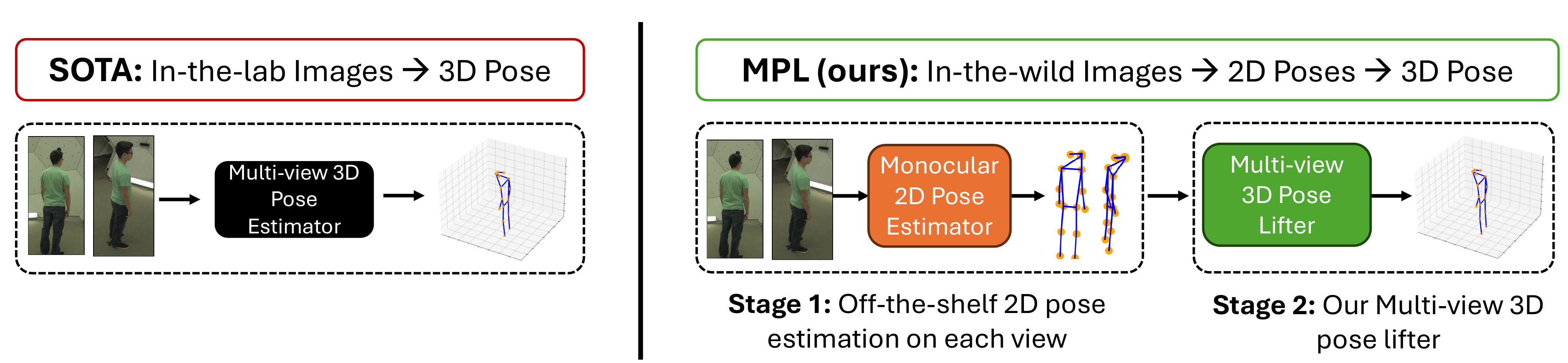}
    \caption{Comparison between the state-of-the-art and our MPL. On the left, the approach used by most prior works. It takes images as input and directly predicts a 3D pose. On the right, our approach splits the process into two stages: 1) Performing 2D pose estimation on each view, and 2) Fusing the information from all views to predict a 3D pose. As a key advantage, our MPL can be trained for arbitrary acquisition setups, using synthetic 2D-3D pairs of skeletons. In contrast, prior methods build on images-3D pose correspondences, which are only available for specific scenes and acquisition conditions, thereby penalizing the generalization capabilities of trained models.}
    \label{fig:pull}
\end{figure*}

The estimation of the position and movement of the human joints and limbs (a.k.a anatomical keypoints) has been a long lasting problem in Computer Vision. Its applications include human tracking and re-identification, but also behavioral analysis in a large variety of fields, \eg movement in sports, fall detection of the elderly, and stroke recovery analysis in neuroscience \cite{Somers_2023_WACV,bridgeman2019multi,salchow2022emerging,el2023systematic,DeepSportLab}.

Deep neural networks have pushed the boundaries of 2D human pose estimation (HPE). In contrast, estimating 3D poses from monocular 2D images remains an ill-posed problem due to the ambiguity inherent to projective geometry. 
Cases including occlusions and oblique viewing are particularly difficult to solve, which motivates the use of a multi-view camera setup \cite{Simon_2017_CVPR, wang2019geometric, h36m, chen2021mvhm} to enrich the available visual information, and boost the performance of both 2D and 3D pose detection \cite{ma2021transfusion, ma2022ppt, wang2021mvp}. 
These works, however, fail to generalize properly to real-world scenarios, also referred to as 'in-the-wild', since their training relies on multi-view images paired with 3D poses, which are scarce and limited to 'in-the-lab' conditions \cite{h36m, cmu}.
This offers limited variability in terms of both acquisition setup and scene appearance, making these solutions inappropriate for practitioners interested in analyzing the 3D pose/gait in real world conditions.
In contrast, our work aims at developing a 3D human pose estimation system suited to 'in-the-wild' deployment, capable of handling arbitrary scenes and acquisition setups.

To alleviate the dependency on multi-view human images paired with 3D poses, we propose decoupling the 3D human pose estimation from the anatomical keypoint detections in 2D views. As shown in \cref{fig:pull}, this is achieved by splitting the 3D pose estimation in two steps. First, 2D skeletons are independently extracted from the available images using an off-the-shelf robust 2D pose estimator. Second, an original 3D pose lifter transforms the (multiple) 2D skeletons into a 3D skeleton.
In the second step, the fusion of different views is implemented in the 2D pose feature space, unlike most prior works that work in the image feature space. The simplest way to make this fusion is to triangulate the anatomical keypoints. This approach is considered as a reference baseline in this paper, but appears to offer sufficient accuracy only when all keypoints are visible and correctly detected in multiple views. 

To mitigate this limitation, we propose to train a transformer-based network called Multi-view 3D Pose Lifter (MPL), which takes 2D keypoint coordinates 
from each view and returns a 3D skeleton in the world coordinate system. 

The most important advantage of decoupling 2D pose estimation and 3D pose lifting is that only pairs of 2D and 3D poses are needed to train our 3D pose lifter, eliminating the need for images paired with 3D poses. 
Creating synthetic 2D-3D pose pairs is much cheaper than generating images-3D pose pairs.
We explain in \cref{ssec:method_dataset_pipeline} how pairs of 2D-3D poses can be generated for arbitrary viewpoints, using the AMASS human shape dataset \cite{AMASS:ICCV:2019}. In particular, the noise associated with 2D pose estimation models is taken into account by deriving the 2D pose from images obtained through 3D mesh rendering tools.
The possibility to generate training set for an arbitrary camera setup makes our method deployable in various acquisition conditions.


The rest of the paper is organized as follows. \cref{sec:related} positions our work with respect to related works. \cref{sec:methodology} details the two main contributions of the paper, namely our proposed 3D pose lifter, and the method designed to derive its 2D-3D pose training set. Finally, experimental validation is presented in \cref{sec:exp}. 

\section{Related Work}
\label{sec:related}
\subsubsection{Monocular 3D Pose Estimation}
The current state-of-the-art in single-view 3D HPE can be categorized into two main categories. 
The first category consists of methods that infer the 3D coordinates directly from images using convolutional neural networks. These methods are complex and, more importantly, limited to the in-the-lab scenes \cite{sun2018integral,pavlakos17volumetric}.

The second category includes methods that use high quality 2D pose estimators and turn the 2D poses into 3D poses with deep neural networks, \eg convolutional, recurrent, and transformers. These methods mostly rely on temporal consistency. 
They are simple, fast, and can be trained on motion capture datasets making them more accurate in real-world scenes compared to the works from the first category \cite{motionbert2022,Shan_2023_ICCV_d3dp,Zhang_2022_CVPR_mix_ste,pose_former_v2:2023}. However, due to working with a single view, they are unable to predict the 3D poses in world coordinates.

\subsubsection{Multi-view 3D Pose Estimation}
In recent years, there has been an increasing interest in dealing with multi-view 3D HPE by fusing the image features from all views \cite{ma2022ppt,ma2021transfusion,epipolartransformers,multiviewpose_cross_view_fusion,iskakov2019learnable}. 
Epipolar Transformers \cite{epipolartransformers} fuses the features of one pixel in one view with features along the corresponding epipolar line of the other views. 
Cross View Fusion \cite{multiviewpose_cross_view_fusion} learns a fixed attention matrix for fusing features in all other views. 
TransFusion \cite{ma2021transfusion} applies transformers globally to fuse features of the reference views. 
Learnable-triangulation \cite{iskakov2019learnable}  aggregates the features from different views in a volume by unprojecting the features using the camera calibration parameters. 
Finally, PPT \cite{ma2022ppt} utilizes a transformer-based approach for 2D pose estimation and makes use of the attention matrix weights to prune the unimportant visual tokens.
However, all these methods require multi-view images paired with annotated 3D pose, making them suited to images corresponding to the scene and acquisition setup used at training but vulnerable to novel deployment conditions.

The closest work to ours is probably \cite{kadkhodamohammadi2021generalizable} which proposes concatenating all the joints' 2D coordinates from all views and treating them as a single input to a fully connected network trained to predict the 3D pose in world coordinates. However, we show that a fully connected network is prone to strong over-fitting and can fall short in accuracy when facing unseen data. 

\section{Methodology}
\label{sec:methodology}
\begin{figure*}[t]
    \centering
    \includegraphics[width=\textwidth,trim={1cm 2cm 1cm 0cm},clip]{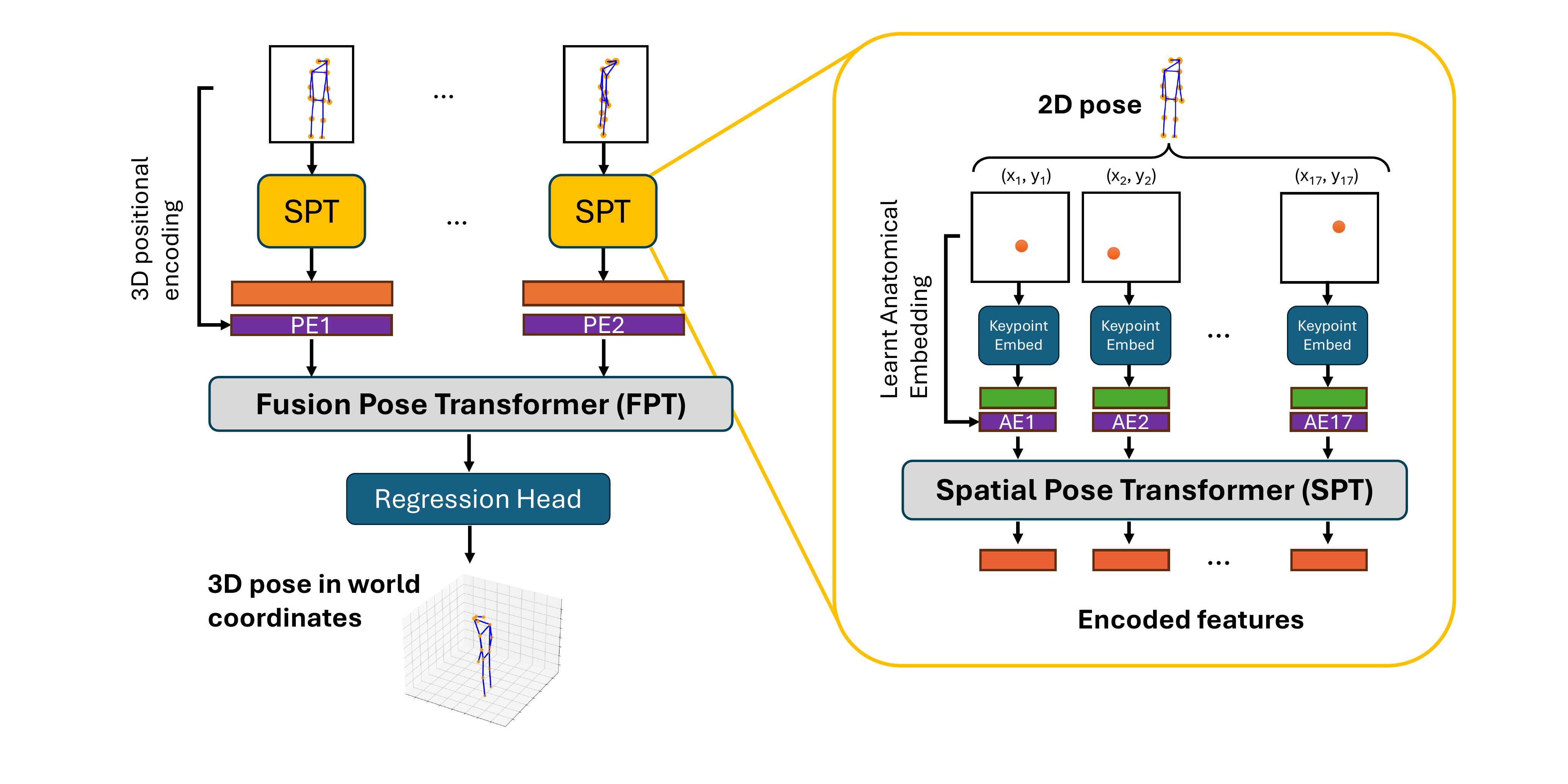}
    \caption{MPL takes 2D poses from all views as inputs, encodes them independently, and fuses them using a transformer network.}
    \label
{fig:overview}
\end{figure*}

As shown in \cref{fig:pull}, our 3D human pose estimation pipeline encompasses two stages. First, the human 2D pose is estimated independently in each view. Second, the 2D poses are lifted to the 3D space, assuming the views are calibrated.
This section introduces the two key contributions of our work, namely our 3D pose lifter (Section \ref{ssec:pose-lifter}) and the approach adopted to generate its training set for an arbitrary acquisition setup (Section \ref{ssec:method_dataset_pipeline}). 

\subsection{Multi-view 3D Pose Lifter}
\label{ssec:pose-lifter}
Our Multi-view Pose Lifter (MPL) for estimating 3D human pose  from the 2D poses observed in a calibrated multi-view system is depicted in \cref{fig:overview}.
MPL takes as input the 2D pose skeletons computed by an off-the-shelf 2D pose estimator in the $N$ views. Let $P_{i} \in \mathbb{R}^{K \times 2}$ denote the $K$ pairs of coordinates defining the location of the $K$ anatomical keypoints in the $i^{th}$ view. MPL takes $\{ P_{i} \}_{i<N}$ as input and predicts the 3D pose skeleton $Q \in \mathbb{R}^{K \times 3}$ in the world coordinate system.
MPL involves two components, namely the Spatial Pose Transformer (\emph{SPT}) and the Fusion Pose Transformer (\emph{FPT}), defined as follows.

\subsubsection{Spatial Pose Transformer (SPT)}

SPT pre-processes each skeleton $P_{i}$ independently of other 2D skeletons.
It is identical for all views and encodes the keypoints in a format suited to the subsequent fusion module.
Therefore, it takes the keypoint coordinates $P_{i} \in \mathbb{R}^{K \times 2}$ as input and returns a $D$-dimensional hidden embedding for each anatomical keypoint. 
First, a linear projection is applied to project each 2D coordinate to a $D$-dimensional vector, resulting into $X_{i}(k) \in \mathbb{R}^{D}$ for the $k^{th}$ keypoint. Then an embedding that is learnt for each type of keypoint is added to each vector of transformed coordinates. This makes the transformer aware of the joints type information \cite{attn_all_need}. Those $K$ $D$-dimensional anatomical embeddings are denoted $AE(k)$, with $1 \leq k \leq K$, and $X^\prime_{i}(k) = X_{i}(k) + AE(k)$ provides the $k^{th}$ keypoint token associated to the $i^{th}$ view. 
Next, the $K$ keypoint tokens associated to a view are fed to an encoder transformer that returns $K$ vectors $X_{i}^S(k) \in \mathbb{R}^{D}$, with $1 \leq k \leq K$, for view $i$.
We adopt the same architecture as in \cite{pose_former:2021} for the encoder transformer layers. It relies on multi-headed self-attention (MHSA) and multi-layer perceptron (MLP). The transformer network has $L$ encoder layers and each layer has $H$ heads \cite{attn_all_need}.


\subsubsection{Fusion Pose Transformer (FPT)}

The fusion transformer takes all keypoint tokens as inputs, namely $X_{i}^S(k)$ for view $i$ and keypoint $k$.
To limit the complexity of the attention module, the keypoint tokens associated with a view are concatenated to obtain a vector $Y(i) \in \mathbb{R}^{K \times D}$, for each view $i$, with $1 \leq i \leq N$.

A 3D positional encodings $E_{3D}(i) \in \mathbb{R}^{K \times D}$, learnt for each view $i$, is first added to each $Y(i)$. This is expected to make the transformer aware of the camera position information, resulting in a view token $Y^\prime(i) = Y(i) + E_{3D}(i)$ for the $i^{th}$ view.
Next, the view tokens are fed to an encoder transformer, which returns a fused embedding $Y^F(i) \in \mathbb{R}^{K \times D}$.
The transformer encoder in FPT follows the same scheme as SPT. 
Eventually, 
$Y^F$ is passed to the regression head to make the final output, \ie, the 3D pose skeleton $Q \in \mathbb{R}^{K \times 3}$.
The regression head is composed of a weighted sum and a final 1-layer MLP to predict the 3D pose in the world coordinates. 
Specifically, the weighted sum takes as input $N$ vectors $Y^F(i)$ corresponding to $N$ views and combines them with a weighted summation whose weights are learnt through training.

\subsubsection{Loss Function}

To train MPL, the output is directly compared to the ground truth 3D pose using Mean Per Joint Position Error (MPJPE).
\cref{eq:mpjpe} shows how MPJPE is calculated, with $Q$ and $\hat{Q}$ denoting the predicted and the ground truth 3D pose, respectively.

\begin{equation}
  \text{MPJPE} = \dfrac{1}{K} \sum_{k=1}^{K} \norm{Q_{k} - \hat{Q_{k}}}^2 .
  \label{eq:mpjpe}
\end{equation}

\subsection{Our Mesh-based 2D-3D Human Pose Dataset Generator}
\label{ssec:method_dataset_pipeline}

\begin{figure}[t]
    \centering
    \includegraphics[trim={0cm 5cm 0cm 5cm},clip,width=\columnwidth]{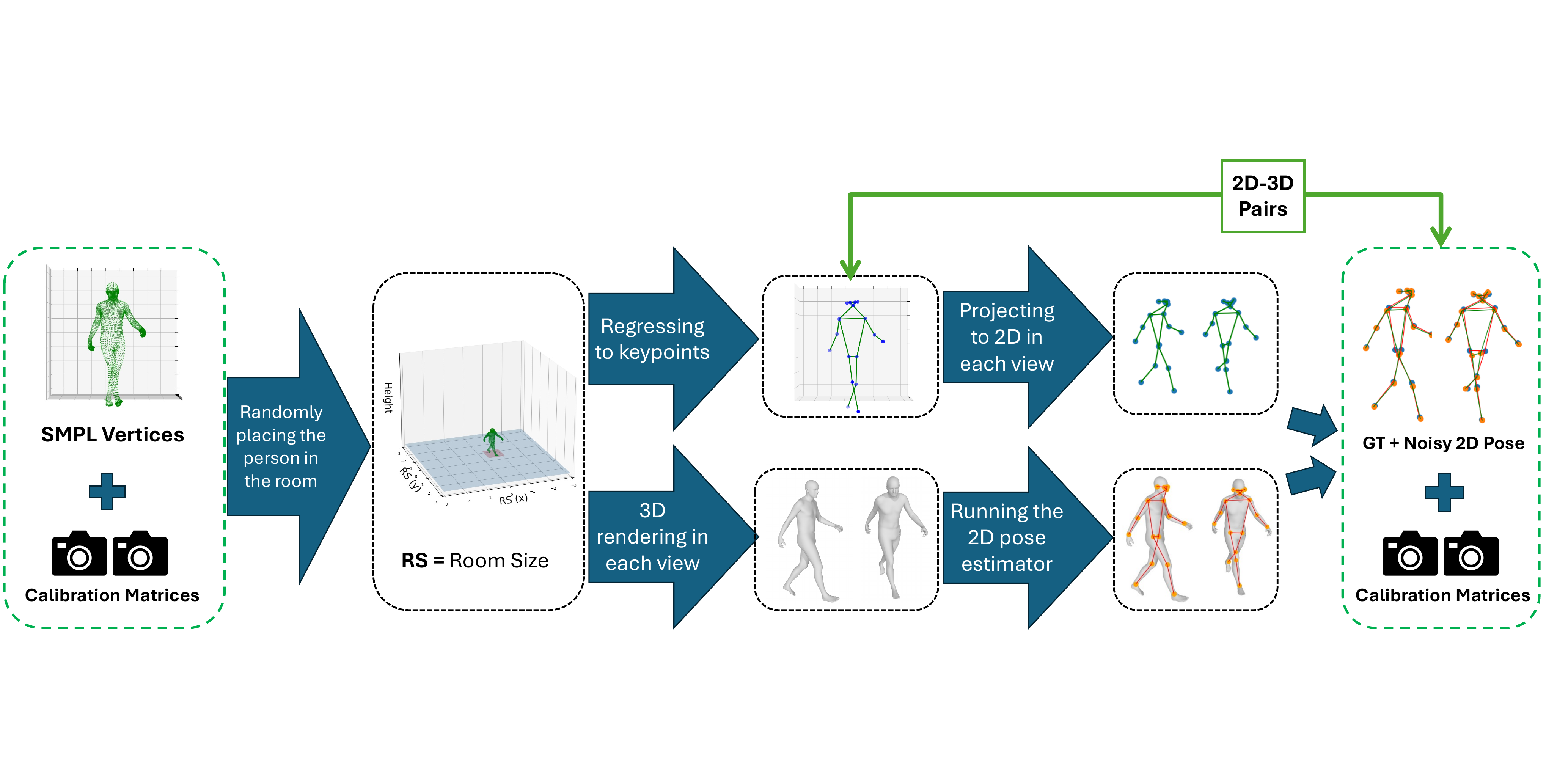}
    \caption{Our \textbf{Mesh-based Human Pose Dataset Generator (MHP)} takes 3D mesh vertices and camera calibration matrices as inputs. It randomly positions the mesh in the scene and returns the 3D pose ground truth and one noisy 2D pose per camera view. The noise associated with the 2D poses comes from the fact that each 2D pose is estimated by an off-the-shelf 2D pose estimation model, applied to the image of the mesh rendered in the corresponding view. The 3D keypoint regressor defines the 3D pose ground truth in a way that is consistent with the definition of ground truth at testing. This makes the 2D-3D pairs of human pose appropriate for training an accurate and robust MPL, able to turn the 2D poses computed by the off-the-shelf pose estimator into a 3D skeleton that is consistent with the 3D keypoints expected at inference.}
    \vspace{-0.6cm}
    \label{fig:dataset_build}
    
\end{figure}

The most natural way to generate the 2D-3D pairs required to train our MPL for a given camera setup with known calibration parameters is to position a 3D skeleton in arbitrary positions within the observed 3D space and project this 3D skeleton into a view of interest, thereby generating the corresponding 2D skeleton.  

This strategy, however, suffers from two important drawbacks. First, it appears that the 3D skeleton datasets adopt distinct rules to define the keypoints. Most often, those definitions are not compatible with the ones adopted by the off-the-shelf 2D pose estimation model used to process 2D images at inference, making the 2D-3D associations considered at training incompatible with the 2D pose manipulated at inference. 
Second, this natural 3D skeleton projection strategy is unable to mimic the noise and errors affecting 2D pose estimation at inference.

To alleviate those issues, instead of considering 3D skeleton and their 2D projections to train our MPL, we propose to generate the 2D-3D pairs of skeletons from meshes.





Therefore, we make use of AMASS \cite{AMASS:ICCV:2019}, an archive of motion capture datasets based on SMPL parameters \cite{SMPL:2015}. This dataset provides a variety of human shapes collected from many datasets and represented with standard 3D mesh models, convertible to SMPL vertices \cite{SMPL:2015} that can be transferred to keypoints by regression \cite{li2021hybrik}.

Considering a dataset of 3D meshes offers two valuable properties. First, the 2D pose associated with a 3D mesh can be created by applying the off-the-shelf 2D pose estimation model used at inference on synthetic 2D renderings of the 3D mesh. This makes the 2D-to-3D pose lifter robust to noise induced by the 2D pose estimator. Second, the regressor used to convert the 3D mesh into a 3D skeleton can be designed to be compatible with the definition of keypoints adopted by the 2D pose estimator.

\cref{fig:dataset_build} shows the pipeline of our proposed Mesh-based Human Pose Dataset Generator (MHP) for 2D-3D pairs of human skeletons generation. It depicts how a dataset is created for a given room size and camera setup using AMASS 3D meshes regressed to a $K$-joint skeleton in the format used by the dataset under evaluation. 
In practice, we render the 3D meshes from AMASS in the camera space of each view based on the camera calibration parameters using the open-source softwares \textbf{Body Visualizer}\cite{SMPL-X:2019} and \textbf{Trimesh} \cite{trimesh}. Then the rendered images are passed through the 2D pose estimator, \ie, the same as the one used at inference, to obtain the 2D poses. Finally, the predicted 2D poses from different views paired with the regressed 3D keypoints are used during training. An overview of this process is shown in \cref{fig:dataset_build}.


\section{Experimental validation}
\label{sec:exp}

\subsection{Evaluation Metric and Datasets}
\label{subsec:eval}

The 3D pose is evaluated by MPJPE (see \cref{eq:mpjpe}), computed in world coordinates \textit{mm}, between the ground truth 3D pose and the predicted 3D pose. 
We evaluate our MPL and alternative methods on two commonly used 3D human pose datasets, Human3.6M \cite{h36m} and CMU Panoptic \cite{cmu}. In all our experiments, no image or 3D pose of the evaluation dataset is used to design or train the tested method.

\subsubsection{Human3.6M} \cite{h36m} is the most widely used public dataset for single-person 3D HPE. It consists of 11 professional actors performing 17 actions such as talking, walking, and sitting. The videos are recorded from 4 different views in an indoor environment. In total, this dataset contains 3.6 million video frames with 3D ground truth annotations, collected by an accurate marker-based motion capture system. Prior works \cite{li2021hybrik, ma2022ppt, pose_former:2021}, utilized all 15 actions for training their models on five subjects (S1, S5, S6, S7, and S8) and tested them on two subjects (S9 and S11). We consider the same two subjects for testing our network.
For evaluating our MPL on Human3.6M, we generate the MPL training 2D-3D pose dataset utilizing MHP from AMASS meshes and the camera calibration parameters of Human3.6M. Thus, we do not use any 3D pose data from Human3.6M for training. 

\begin{figure*}[t]
    \centering
    \begin{subfigure}[b]{0.5\textwidth}
         \centering
         \includegraphics[trim={3cm 3cm 2.5cm 3cm},clip,width=\textwidth]{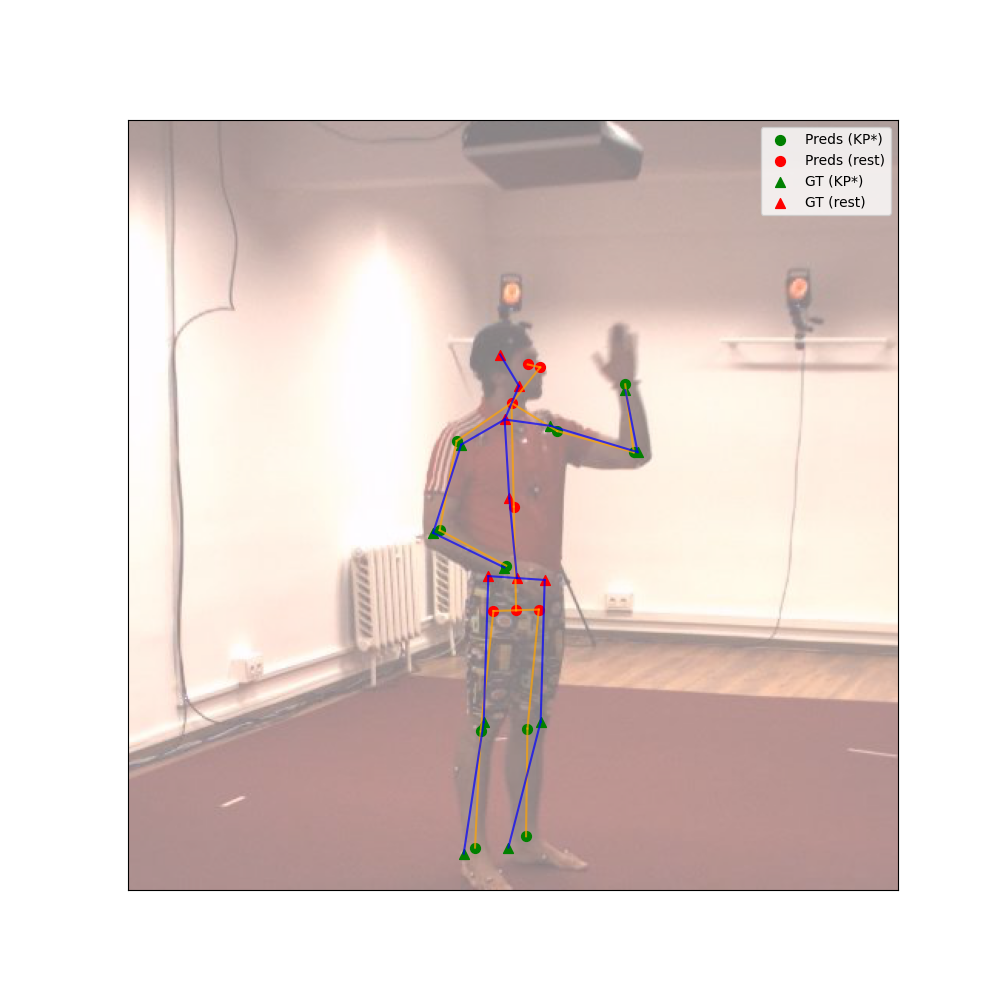}
         \label{subfig:vis}
     \end{subfigure}
    
    \caption{Example of inconsistency between Human3.6M and COCO format. The triangles depict the 3D ground truth keypoints projected to 2D space (Human3.6M format), and the circles depict the keypoints predicted by the off-the-shelf 2D pose estimator (COCO format). The green keypoints are the ones that are similarly defined in both formats, and the red ones correspond to the keypoints that have distinct definitions in the two formats. Note that, some of the predicted keypoints are estimated by averaging the locations of a subset of keypoints predicted by the 2D pose estimator (in COCO format), \eg the keypoint \textit{torso} is located in the middle of \textit{pelvis} and \textit{neck}. We do not include these averaged keypoints in the set of keypoints KP*, corresponding to keypoints with similar definition, even if their location is reasonably similar to the one adopted in Human3.6M.}
    \vspace{-.3cm}
    \label{fig:inconsist-format}

\end{figure*}

It is worth noting that the definition of keypoints in the Human3.6M dataset differs from the one introduced by COCO \cite{coco:Lin2014MicrosoftCC} and adopted by the off-the-shelf 2D pose estimator used in the study. 
By averaging the location of some COCO keypoints, we have improved the consistency between the two formats. For instance, averaging the \textit{ears} and \textit{eyes} locations from the COCO format results in a keypoint that closely matches the \textit{head} keypoint in the Human3.6M format. Similarly, averaging the \textit{hips} in COCO provides the \textit{pelvis} in Human3.6M, and the mean of the \textit{neck} and the newly created \textit{pelvis} results in a \textit{torso} keypoint.
\cref{fig:inconsist-format} illustrates this process. The set of keypoints that adopt consistent definitions in both formats is denoted KP* and includes the knees, the ankles, the shoulders, the elbows, and the wrists.

\subsubsection{CMU Panoptic} \cite{cmu} is a more recent multi-view public dataset made available by the Carnegie Mellon University. The dataset contains Full-HD video streams of 40 subjects captured/observed from up to 31 cameras located in a dome. 3D ground-truth pose annotations are generated via triangulation using all camera views. Following previous works \cite{iskakov2019learnable, Xiang2018MonocularTC}, we use the 17-keypoint subset (from the 19-joint annotation format) that are in line with the popular COCO format \cite{coco:Lin2014MicrosoftCC}. 

For evaluating on CMU, we generate the MPL training 3D-2D pose dataset utilizing our MHP with AMASS meshes and the CMU camera calibration parameters. Note that the CMU motion capture that is part of the AMASS collection has not been included in any of our training datasets. Finally, we test our MPL on the test split recommended in \cite{Xiang2018MonocularTC}, where only single-person scenes are selected.

\subsection{2D Pose Prediction and MPL Implementation Details}
\label{sec:implementation}


For predicting the 2D poses in the image space, we use the open-source pose estimator framework \textbf{\mmpose} \cite{mmpose2020}. 
It involves the popular HRNet \cite{sun2019deep} architecture, pretrained on COCO
\cite{coco:Lin2014MicrosoftCC}.
The average precision (AP) \cite{coco:Lin2014MicrosoftCC} of the \mmpose 2D output poses on AMASS rendered images are 95.9\% and 93.2\% when applied with Human3.6M and CMU acquisition setups, respectively.
Note that AP is computed when Object Keypoint Similarity (OKS) \cite{coco:Lin2014MicrosoftCC} threshold is set to 0.75 \footnote{Please refer to \cite{coco:Lin2014MicrosoftCC} for more information about OKS and AP.}. This shows that the chosen off-the-shelf 2D pose estimator 
is able to effectively handle 3D rendered human mesh images.

We set $K$, \ie, the number of joints, to 17 in our configurations. The hidden embeddings dimension $D$ is set to 32 in our MPL. Both SPT and FPT have the same configuration parameters. 
The number of encoder heads $H$ is set to 8, and the number of encoder layers $L$ is set to 2 for the experiments on CMU and 12 for the experiments on Human3.6M.
During MPL training, the input batch size is set to 32, and the network is optimized using the Adam optimizer \cite{Kingma2014AdamAM} with MPJPE loss, for 20 epochs. The learning rate is initialized at 0.0001, and decays at the 10th and 15th epochs with a decreasing ratio of 0.1.

The training dataset is created separately for each evaluation using our proposed MHP. For the evaluation of Human3.6M and CMU, the MHP's room size is set to $2 \times 3.5$ and $0.4 \times 0.4$ meters, respectively. In each case, the corresponding camera calibration information is used.


\subsection{Comparison to Baselines and Prior Works}
\label{sec:results}

Since our contribution primarily aims to enable 3D pose estimation in arbitrary observation conditions, a fair positioning should compare our MPL to works that are either not trained (typically because they use triangulation on top of 2D pose estimation) or trained on other datasets than the one used at testing. 
Trained alternatives with publicly available code and trained weights are rare among previous works. We have identified three methods, namely Learnable\hyp{}triangulation \cite{iskakov2019learnable}, TransFusion \cite{ma2021transfusion}, and PPT \cite{ma2022ppt}. 

Learnable\hyp{}triangulation \cite{iskakov2019learnable} aggregates features extracted from the image \linebreak 
spaces by unprojecting them into a common voxel space. The publicly available data only provides the trained weights on Human3.6M; thus we test their network only on CMU dataset. 
TransFusion \cite{ma2021transfusion} trains a transformer\hyp{}based network that, by globally attending features in the image space, fuses the information from different views. The trained weights are only provided on Human3.6M so we only test their network on CMU.
Finally, PPT \cite{ma2022ppt} follows the same approach as TransFusion but, instead of considering a global attention process in the fusion process, it prunes the unimportant tokens. Trained weights are only provided on Human3.6M. 
However, since PPT is more recent and provides more accurate results than TransFusion \cite{ma2022ppt}, we follow their guidelines to train their network on CMU dataset and also test on Human3.6M.

In addition, we compare our MPL to a 3D pose lifter implemented based on a Fully Connected network, following the guidelines in \cite{kadkhodamohammadi2021generalizable}, instead of our transformer\hyp{}based solution.



\subsubsection{Human3.6M}



\cref{table:mpjpe-h36m} shows the 3D HPE results on the test set of Human3.6M for each action. 
To evaluate the impact of the mismatch between the COCO and Human3.6M formats (see Section \ref{subsec:eval}) on the evaluation metrics, the last column in \cref{table:mpjpe-h36m} computes the MPJPE only on the keypoints whose definitions are consistent in both the 3D and 2D formats. This set of keypoints is denoted KP* and corresponds to knees, ankles, shoulders, elbows, and wrists.

The results show that PPT \cite{ma2022ppt} is unable to generalize well to unseen images. Even with a higher number of views, it falls short both compared to the triangulation baselines and to our MPL. 
This explains why triangulation is the reference when 3D pose estimation has to be deployed 'in-the-wild'. 

Utilizing MPL improves the MPJPE (KP*) on Human3.6M test set by 52.6 $mm$ on average when using two views. This corresponds to an MPJPE reduction of about $45\%$. This gain raises to $56\%$ when accounting for keypoints for which the 3D ground truth definition is not consistent with the one adopted by the 2D pose estimator. This reveals the ability of our MPL to predict 3D keypoints that (moderately) differ from the keypoints predicted by the off-the-shelf 2D pose estimator, thereby increasing the flexibility of our framework\footnote{For example, 3D keypoints could be defined as internal body points, even if the associated 2D keypoints are defined based on visual, but 3D-inconsistent, cues in the projected views.}. 

As expected, \cref{table:mpjpe-h36m} shows that the enhancement obtained with MPL over triangulation decreases when increasing the number of views. This is because when increasing the number of camera views, there is more chance of obtaining correct 2D poses from, at least, two views. Hence, no need for a more sophisticated method than triangulation.
Moreover, \cref{table:mpjpe-h36m} shows that replacing the MPL transformer by a fully connected network (leading to a solution close to \cite{kadkhodamohammadi2021generalizable}) results in a significantly worse pose lifting.

\begin{table*}[t]
\centering
\caption{3D MPJPE in $mm$ on the Human3.6M test set for different methods. The last but one column averages the MPJPE on the multiple actions considered in the test set. The last column does the same but only considers the subset of keypoints KP* for which the 3D ground truth definition (adopted by Human3.6M) is consistent with the definition adopted by the \mmpose 2D pose estimator. The numbers in bold indicate the lowest MPJPE with two views.}
\label{table:mpjpe-h36m}
\resizebox{\textwidth}{!}{
\begin{tabular}{l|c|*{15}{P{1.2cm}}|c c}  
\toprule
\multirow{3}{*}{Method} & \multirow{3}{*}{\begin{tabular}[c]{@{}c@{}}\#\\ Views\end{tabular}} & \multirow{3}{*}{Direction} & \multirow{3}{*}{Discuss} & \multirow{3}{*}{Eating} & \multirow{3}{*}{Greet} & \multirow{3}{*}{Phone} & \multirow{3}{*}{Photo} & \multirow{3}{*}{Pose} & \multirow{3}{*}{Purchase} & \multirow{3}{*}{Sitting} & \multirow{3}{*}{\begin{tabular}[c]{@{}c@{}}Sitting\\ Down\end{tabular}} & \multirow{3}{*}{Smoke} & \multirow{3}{*}{Wait} & \multirow{3}{*}{\begin{tabular}[c]{@{}c@{}}Walk\\ Dog\end{tabular}} & \multirow{3}{*}{Walk} & \multirow{3}{*}{\begin{tabular}[c]{@{}c@{}}Walk\\ Two\end{tabular}} & \multicolumn{2}{c}{Avg.} \\ 
\cmidrule{18-19}
& & & & & & & & & & & & & & & & & (All KP) & (KP*) \\
\midrule
PPT \cite{ma2022ppt} & 4 & 677.5 & 162.8 & 115.0 & 209.3 & 136.7 & 286.6 & 177.4 & 226.9 & 253.3 & 218.6 & 179.2 & 99.7 & 81.2 & 155.6 & 98.3 & 196.2 & 274.41 \\
\midrule
\multirow{2}{*}{Triangulation} & 2 & 157.4 & 121.4 & 81.8 & 149.3 & 130.3 & 105.5 & 101.7 & 201.6 & 98.4 & 114.4 & 107.5 & 175.8 & 73.1 & 88.6 & 80.0 & 121.2 & 114.7 \\
& 4 & 72.3 & 60.4 & 51.6 & 58.6 & 84.4 & 55.2 & 59.0 & 66.1 & 56.8 & 55.9 & 59.6 & 53.4 & 48.4 & 59.3 & 50.1 & 60.6 & 49.2 \\
\midrule
\multirow{3}{*}{FullyConnected} & 1 & 142.8 & 144.8 & 168.8 & 146.5 & 227.0 & 131.4 & 145.1 & 339.8 & 253.2 & 185.5 & 169.5 & 163.5 & 172.2 & 172.4 & 168.3 & 186.3 & 195.8 \\
& 2 & 108.6 & 102.8 & 115.4 & 110.9 & 131.2 & 98.5 & 100.6 & 166.2 & 165.2 & 121.8 & 119.9 & 111.1 & 108.1 & 107.6 & 106.0 & 119.4 & 131.6 \\
& 4 & 64.6 & 77.6 & 72.5 & 69.3 & 116.4 & 65.0 & 73.7 & 181.1 & 111.4 & 91.2 & 83.7 & 76.7 & 76.8 & 72.7 & 77.6 & 90.7 & 102.4 \\
\midrule
\multirow{3}{*}{\textbf{MPL} (ours)} & 1 & 135.6 & 131.0 & 141.2 & 124.2 & 190.1 & 118.7 & 120.0 & 303.5 & 200.0 & 168.3 & 151.8 & 133.0 & 143.7 & 144.8 & 141.4 & 161.0 & 169.5 \\
& 2 & \textbf{46.6} & \textbf{47.8} & \textbf{48.3} & \textbf{44.4} & \textbf{64.5} & \textbf{42.8} & \textbf{46.8} & \textbf{76.3} & \textbf{68.8} & \textbf{57.2} & \textbf{50.3} & \textbf{46.4} & \textbf{47.0} & \textbf{48.6} & \textbf{45.9} & \textbf{53.3} & \textbf{62.1} \\
& 4 & 33.2 & 33.8 & 34.7 & 31.9 & 40.4 & 32.0 & 34.8 & 45.8 & 50.9 & 38.8 & 35.5 & 32.0 & 35.4 & 34.8 & 33.8 & 36.9 & 42.9 \\ 
\bottomrule
\end{tabular}
}
\end{table*}


\subsubsection{CMU Panoptic}

We follow \cite{voxelpose, wang2021mvp} and evaluate our MPL on the CMU dataset based on a subset of HD cameras (3, 6, 12, 13, 23). \cref{table:mpjpe-cmu} shows the MPJPE in $mm$ for different methods. In all methods, none of the images or 3D poses of the CMU dataset have been used at training.

The results show that the off-the-shelf pre-trained methods all fail to generalize to the images in the CMU test set, even when enjoying higher number of views. 
\cref{table:mpjpe-cmu} confirms the observations of \cref{table:mpjpe-h36m} that the triangulation baseline does a good job when increasing the number of views to 5. However, using triangulation with only 2 views increases the MPJPE by about $30\%$ compared to our MPL.

\begin{table*}[t]
\centering
\caption{The comparison of 3D MPJPE on CMU Panoptic test set between different methods in $mm$. Note that the MPJPE (KP*) indicates that the calculations were done only considering the keypoints most aligned with COCO as a more fair comparison. The numbers in bold indicate lowest MPJPE with two views.}
\label{table:mpjpe-cmu}
\resizebox{\textwidth}{!}{
\begin{tabular}{l | c c c | m{.7cm}<{\centering} m{.7cm}<{\centering} | m{1cm}<{\centering} m{1cm}<{\centering} m{1cm}<{\centering} | m{1cm}<{\centering} m{1cm}<{\centering} m{1cm}<{\centering}}
\toprule
Method & \begin{tabular}{@{}c@{}}Learn.-triang. \\ \cite{iskakov2019learnable}\end{tabular} & \begin{tabular}{@{}c@{}}PPT \\ \cite{ma2022ppt}\end{tabular} & \begin{tabular}{@{}c@{}}TransFusion \\ \cite{ma2021transfusion}\end{tabular} & \multicolumn{2}{c|}{Triangulation} & \multicolumn{3}{c|}{ \begin{tabular}{@{}c@{}}Fully \\ Connected\end{tabular}} & \multicolumn{3}{c}{\begin{tabular}{@{}c@{}}\textbf{MPL} \\ (ours)\end{tabular}} \\ 
\midrule
\# Views & 5 & 5 & 4 & 2 & 5 & 1 & 2 & 5 & 1 & 2 & 5 \\
\midrule
$\downarrow$ MPJPE (All KP) & - & 114.2 & 764.8 & 43.0 & 21.7 & 87.1 & 51.3 & 47.4 & 77.6 & \textbf{29.3} & 21.1 \\
$\downarrow$ MPJPE (KP*)  & 130.9 & 108.3 & 1166.2 & 44.0 & 21.8 & 96.0 & 60.1 & 56.2 & 84.4 & \textbf{34.2} & 22.3 \\
\bottomrule
\end{tabular}
}
\vspace{-0.2cm}
\end{table*}

\subsubsection{Inference speed}

To figure out whether our MPL can be a viable alternative to triangulation, it is worth considering its run time.
\cref{table:speed} shows the inference speed of MPL in frames/second where each frame is composed of all the input camera views of a single person. 
The numbers are measured on an NVidia A10 with a 32-core Intel Xeon Gold 6346 CPU running at 3.10GHz.
As shown in the table, MPL is able to run in real-time even when 5 views are used with the minimum batch size. Note that, while increasing the batch size improves the speed, it imposes a constant delay to the output stream. 

\begin{table}[h]
\centering
\caption{The inference speed (frames/second) of MPL for different number of views and batch sizes.}
\resizebox{0.55\textwidth}{!}{
\begin{tabular}{>{\centering\arraybackslash}p{1cm} | >{\centering\arraybackslash}p{2cm} >{\centering\arraybackslash}p{2cm} >{\centering\arraybackslash}p{2cm} >{\centering\arraybackslash}p{2cm}}
\toprule
\multirow{2}{*}{\begin{tabular}{@{}c@{}}\# \\ Views\end{tabular}}
 & \multicolumn{4}{c}{Batch Size} \\
\cmidrule(lr){2-5}
  & 1 & 2 & 4 & 8 \\
\midrule
2 & 287.8 & 575.4 & 1127.1 & 2245.6 \\
3 & 240.9 & 482.0 & 956.4 & 1892.3 \\
4 & 203.4 & 404.9 & 791.4 & 1571.4 \\
5 & 173.4 & 352.1 & 698.5 & 1391.0 \\
\bottomrule
\end{tabular}
}
\label{table:speed}
\end{table}

\vspace{-0.3cm}

\subsubsection{Qualitative Results}
\begin{figure*}[t]
    \centering
    \begin{subfigure}[b]{0.75\textwidth}
         \centering
         \includegraphics[trim={1cm .5cm .5cm 3cm},clip,width=\textwidth]{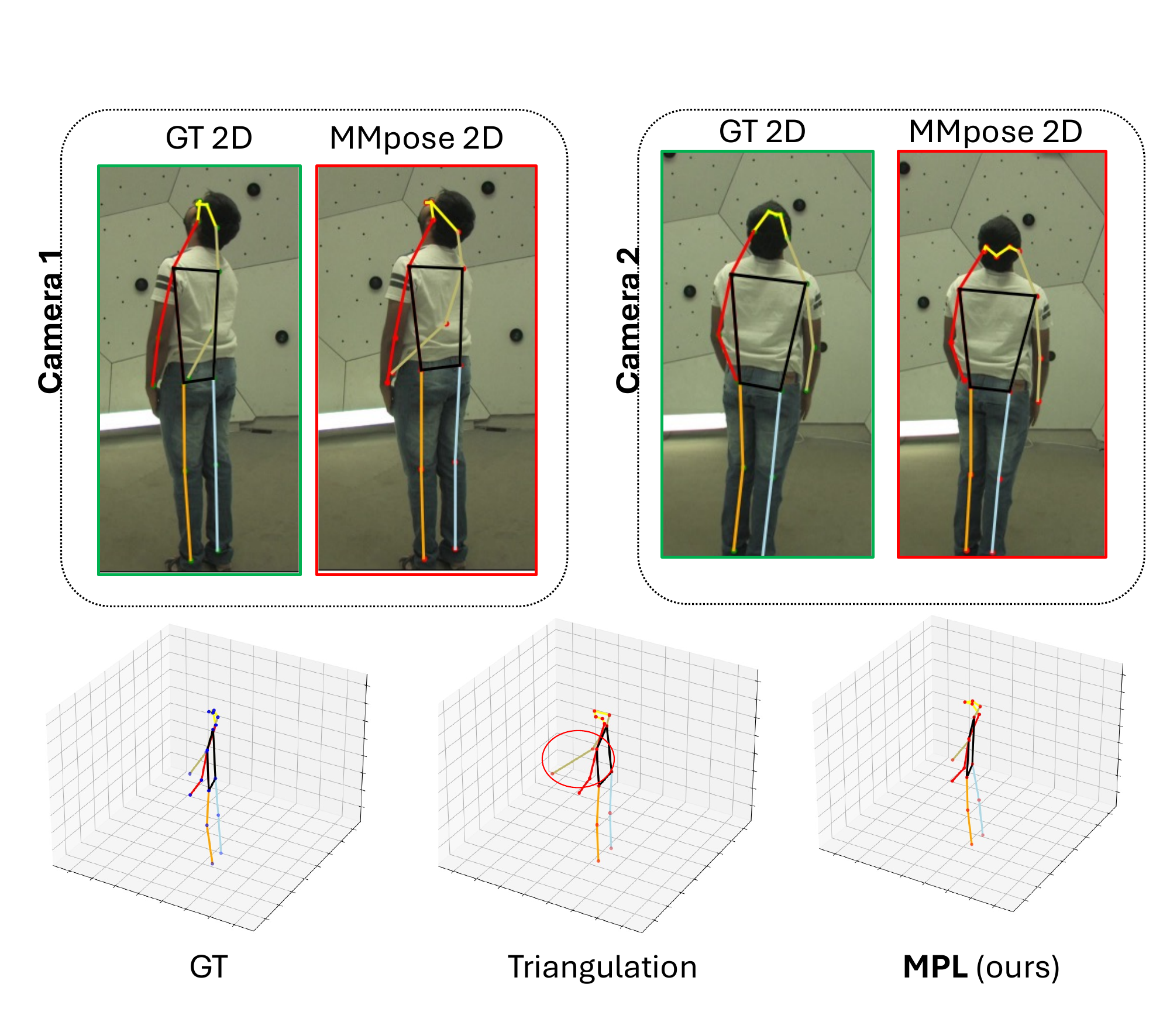}
     \end{subfigure}

    
    \caption{\textbf{Qualitative results.} In the top row, the GT and \mmpose 2D represent the ground truth taken from the CMU dataset and the predicted 2D pose from the off-the-shelf 2D pose estimator, respectively. In the bottom row, the GT shows the 3D ground truth from the CMU dataset, and the other two 3D poses correspond to the outputs of triangulation and MPL when the top row is given as the inputs.}
    \label{fig:qualitative}
\end{figure*}

\cref{fig:qualitative} shows some qualitative results comparing our MPL with triangulation when methods are tested on CMU.

\subsection{Relevance of Our Mesh-based 2D-3D Human Pose Generator}

To show the importance of utilizing our MHP to generate the MPL training set, we consider 3 different scenarios for training our MPL: 
A) training on the 3D and 2D-projected poses of Human3.6M (or CMU),
B) regressing the 3D poses from AMASS meshes, and training based on their corresponding projected 2D poses, and 
C) adopting our proposed MHP, \ie, regressing the 3D pose from AMASS and generating noisy 2D by running the 2D pose estimator on 2D mesh rendering.

\cref{table:importance-3dg} compares the 3D MPJPE of each scenario, when testing on CMU (on the left) and Human3.6M (on the right). We also include the baseline, \ie, triangulation, and only consider the keypoints in KP*, \ie, the keypoints for which the COCO’s format (same as the 2D pose estimator format in our study) is consistent with that of Human3.6M. The number of views for all scenarios is set to two, and the MPJPE is averaged on all possible pairs of cameras, selected among the ones that are commonly used in previous works, \ie, for CMU: HD cameras (3, 6, 12, 13, 23) and for Human3.6M: all four available cameras.

By comparing the Scenario A and B, we observe a $10\%$ drop in MPJPE on both datasets, showing a slight benefit of using AMASS over Human3.6M and CMU. 
Interestingly, the use of our 2D-3D MHP pairs appears to significantly increase the drop in MPJPE, up to $40\%$ compared to pairs derived from the datasets ground truths, thereby revealing that MPH is effective in enriching the training set.

\begin{table}[t]
\centering
\caption{\textbf{The importance of MHP.} MPJPE values are reported in $mm$, the number of views is fixed to two. The testing dataset is CMU and Human3.6M for the table on the left and right, respectively.}
\begin{tabular}{cc}
\begin{minipage}[t]{0.49\textwidth}
\vspace{0pt}
\centering
\resizebox{0.9\textwidth}{!}{
\begin{tabular}{c | c | c c | c }
\multicolumn{5}{c}{\cellcolor{blue!10}On \textbf{CMU}} \\
\toprule
\multirow{2}{*}{Method} & \multirow{2}{*}{Scenario}  & \multicolumn{1}{p{2cm}} {\centering Training \\ Dataset} & \multirow{2}{*}{MHP} & \multirow{2}{*}{\centering $\downarrow$ MPJPE}\\ 
\midrule
Triangulation  & - & - & - & 44.0 \\
\midrule
\midrule
\multirow{3}{*}{\textbf{MPL} (ours)} 
& A & Human3.6M & \xmark & 49.6 \\
& B & AMASS & \xmark & 44.9 \\
& C & AMASS & \cmark & \textbf{34.5} \\
\bottomrule
\end{tabular}
}
\label{table:importance-3dg}
\end{minipage}
&
\begin{minipage}[t]{0.49\textwidth}
\vspace{0pt}
\centering

\resizebox{0.9\textwidth}{!}{
\begin{tabular}{c | c | c c | c }
\multicolumn{5}{c}{\cellcolor{blue!10}On \textbf{Human3.6M}} \\
\toprule
\multirow{2}{*}{Method} & \multirow{2}{*}{Scenario}  & \multicolumn{1}{p{2cm}} {\centering Training \\ Dataset} & \multirow{2}{*}{MHP} & \multirow{2}{*}{\centering $\downarrow$ MPJPE}\\ 
\midrule
Triangulation & - & - & - & 114.7 \\
\midrule
\midrule
\multirow{3}{*}{\textbf{MPL} (ours)} 
& A  & CMU & \xmark & 104.0 \\
& B  & AMASS & \xmark & 94.8 \\
& C & AMASS & \cmark & \textbf{62.1} \\
\bottomrule
\end{tabular}
}
\label{table:importance-3dg}
\end{minipage}
\end{tabular}

\end{table}

\subsection{Impact of Keypoints' Visibility}
\label{sec:visibility}

In order to triangulate a 3D point from corresponding 2D points, you need the 2D points to be visible in, at least, two views. 
This might not be always the case assuming that the cameras are fixed.
This section investigates the impact of keypoint’s visibility on the final 3D error. 
The CMU dataset is chosen for this study because its camera setup offers a wider variety of view angles. We consider a 2-view camera setup, and run our test on all possible pairs selected among the 29 HD cameras available in CMU \footnote{Note that 2 out of the 31 HD cameras in CMU are defected and not considered for this study.}.

\cref{tab:visibility-impact} compares the impact of keypoints’ visibility on their 3D error for both triangulation and our MPL. Here, for each input frame, the keypoint is counted as being visible if and only if it is included in both input images (the fact of being occluded or not has no impact on the visibility count). For each keypoint, the visibility percentage shown in \cref{tab:visibility-impact} is computed over all input frames of all runs. 
Since most keypoints except the ankles and knees are visible most of the time, we only consider the knees and ankles in \cref{tab:visibility-impact}. It shows a 9.7 $mm$ error reduction (for the knees and ankles) when using MPL over triangulation when these keypoints are visible in both views. This reduction rises to 215.4 $mm$ when the keypoints are invisible in at least one of the views, thereby demonstrating the benefit of our trained MPL on invisible keypoints.

\begin{table*}[t]
    \centering
\caption{The comparison of the impact of keypoints' visibility on their 3D reconstruction error for triangulation and our MPL. The number of views is set to two, and we consider all possible combinations of a 2-view setup out of the 29 HD cameras in CMU. The numbers are an average over all cases tested. Note that we only consider keypoints whose visibility (in total) is less than 99\%.}
    
\resizebox{0.95\textwidth}{!}{

\begin{tabular}{c c|c c| m{1.5cm}<{\centering} | m{1.5cm}<{\centering} | m{1.5cm}<{\centering} | m{1.5cm}<{\centering} | m{1.5cm}<{\centering} | m{1.5cm}<{\centering} | m{1.5cm}<{\centering} | m{1.5cm}<{\centering}}
    \toprule
    \multicolumn{4}{c|}{Keypoint} & \multicolumn{2}{c|}{Left Knee} & \multicolumn{2}{c|}{Right Knee} & \multicolumn{2}{c|}{Left Ankle} & \multicolumn{2}{c}{Right Ankle} \\
    \midrule

    \multirow{4}{*}{Visibility} & \multirow{2}{*}{Visible (\%)} & \multirow{4}{*}{Error ($mm$)} & Triangulation & \multirow{2}{*}{75.6} &  57.0 &
    \multirow{2}{*}{80.6} &  60.8 &
    \multirow{2}{*}{35.0} &  59.8 &
    \multirow{2}{*}{35.4} &  60.8 \\
    &&& \textbf{MPL} (ours) & & 45.8 & & 45.6 && 52.8 && 55.2 \\
    \cline{4-12}
    
    & \multirow{2}{*}{Invisible (\%)} & & Triangulation & \multirow{2}{*}{24.4} &  155.9 &
    \multirow{2}{*}{19.4} &  148.6 &
    \multirow{2}{*}{65.0} &  670.7 &
    \multirow{2}{*}{64.6} &  487.9 \\
    &&& \textbf{MPL} (ours) & & 70.1 & & 81.7 && 263.6 && 186.0 \\
    \bottomrule
\end{tabular}
}
\label{tab:visibility-impact}

\vspace{-.3cm}
\end{table*}


This conclusion is confirmed by analyzing the impact of the MHP's parameter room size on the keypoints' visibility and MPJPE. Please refer to the supplementary materials for more details regarding this experiment.

\subsection{Execution Time of Data Preparation and Training}

MPL needs to be trained with data prepared by MHP on a specific camera setup. 
The execution time of MHP depends on the number of views in a camera setup and the number of 3D meshes used from AMASS dataset. For the experiments in this paper, we make use of 128,109 meshes from AMASS. 
It approximately takes 16 hours to run MHP on all 31 HD cameras of CMU on 4 NVidia A10 with 32-core Intel Xeon Gold 6346 CPU running at 3.10GHz.
This number drops to 2 hours for 4 cameras of Human3.6M.

The MPL training time depends on the number of views in the camera setup and the number of layers and heads used in the transformer network. For a 2 and 5-view setup on CMU, with the configuration mentioned in \cref{sec:implementation}, it takes about 26 and 49 minutes, respectively, to train MPL on data prepared by MHP. On Human3.6M with a 2 and 4-view setup, the MPL training takes about 59 and 129 minutes due to using a higher number of layers in the transformer. The training measurements are done on the same device using only 1 GPU.

\section{Conclusion}
We introduce a novel method for estimating 3D human pose from one or multiple calibrated views.
Our research addresses the limitations of the existing methods for multi-view 3D human pose estimation, particularly their dependency on multi-view images paired with 3D poses, which are often limited to controlled, 'in-the-lab' environments. 
By decoupling the process into two stages, being 2D pose estimation and 3D pose lifting, our approach overcomes these constraints and enables more robust performance in real-world scenarios. We leverage the more accurate and widely available 2D pose estimators for the first stage and introduce a transformer-based network, the Multi-view 3D Pose Lifter (MPL), for the second stage.
This network fuses 2D keypoints from multiple views to predict 3D poses in the world coordinates, and can be trained from synthetic pairs of 2D-3D poses, without requiring a set of (generally unavailable) scene-specific images paired with the 3D human poses.
Additionally, we propose a novel pipeline for generating pairs of 2D-3D poses for arbitrary viewpoints with the ability to account for the noise associated to the 2D pose estimation model making our work even more robust.
Our experimental results and ablation studies confirm the effectiveness and versatility of our proposed pipeline, marking a significant step forward in the field of 3D human pose estimation.

\section*{Acknowledgments}

S.A. Ghasemzadeh is funded by the FRIA/FNRS. 
C. De Vleeschouwer is a Research Director of the Fonds de la Recherche Scientifique - FNRS. 
Computational resources have been provided both by the super computing facilities of the Université catholique de Louvain (CISM/UCL) and the Consortium des Équipements de Calcul Intensif en Fédération Wallonie Bruxelles (CÉCI) funded by the Fond de la Recherche Scientifique de Belgique (F.R.S.-FNRS) under convention 2.5020.11 and by the Walloon Region.

%
%
\bibliographystyle{splncs04}
\bibliography{main}

\clearpage
\section*{MPL: Lifting 3D Human Pose from Multi-view 2D Poses: Supplementary Materials}

In the section below, we provide additional details regarding our novel multi-view 3D pose estimator and the mesh-based 3D dataset generator. Our code is available at \url{https://github.com/aghasemzadeh/OpenMPL}.


\subsubsection{The Impact of Person's Range of Displacement.} 
\begin{figure*}[t]
    \centering
    \begin{subfigure}[b]{0.45\textwidth}
         \centering
         \includegraphics[width=\textwidth]{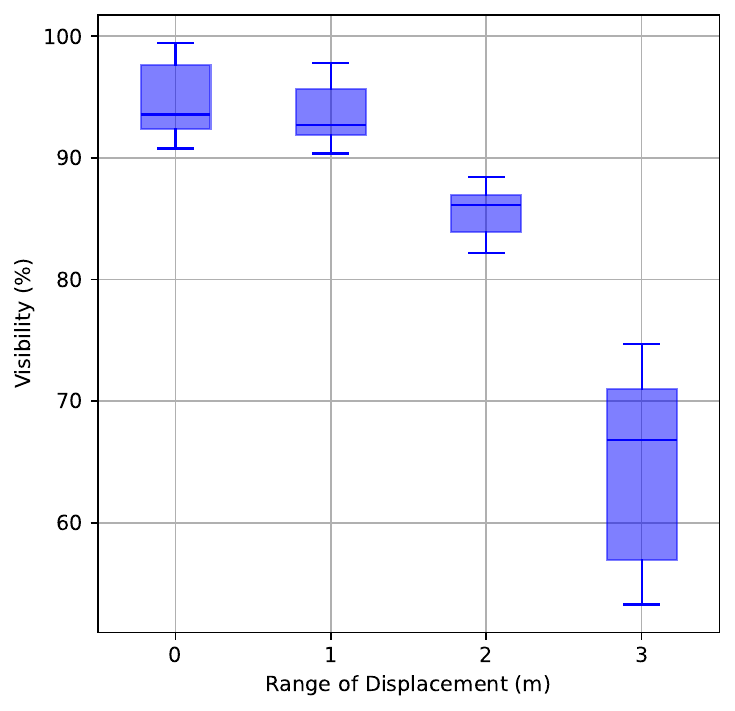}
         \caption{Keypoints' visibility}
         \label{subfig:vis}
     \end{subfigure}
     \hspace{.2cm}
     \begin{subfigure}[b]{0.45\textwidth}
         \centering
         \includegraphics[width=\textwidth]{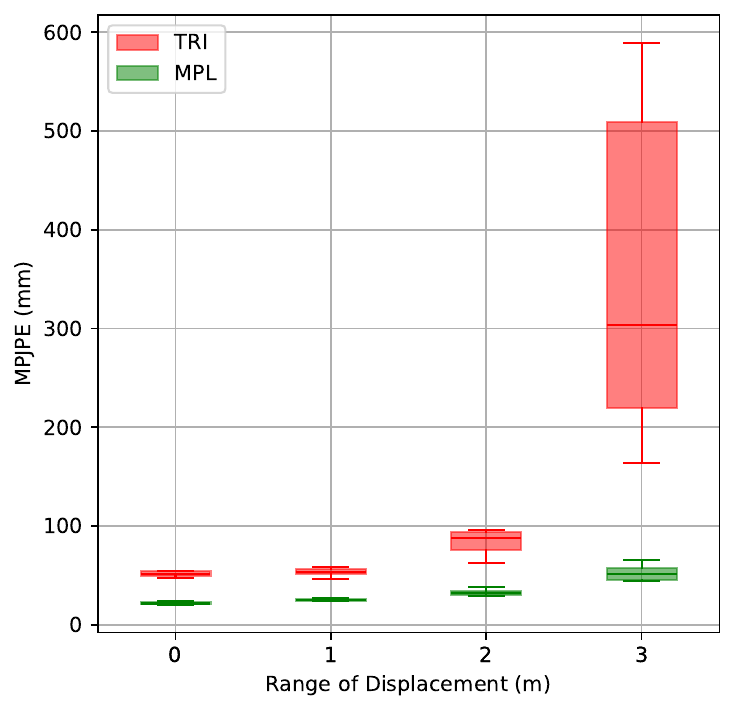}
         \caption{MPJPE}
         \label{subfig:mpjpe}
     \end{subfigure}
    
    \caption{Visibility of keypoints (left) and MPJPE (right) for triangulation and MPL, as a function of the range of displacements adopted to position AMASS human meshes in the scene covered by a fixed camera setup (here defined by the calibration parameters of the two views from the CMU dataset).
    We consider all possible combinations of a 2-view setup out of the 5 selected HD cameras in CMU. Without any surprise, we observe that less keypoints remain visible in both views when the range of displacement increases. This translates to a more severe MPJPE penalty for triangulation than for our MPL, which appears to be robust to missing keypoints.}
    \vspace{-.4cm}
    \label{fig:amass-range-of-freedom}
\end{figure*}

Obviously, a way to reduce the keypoint visibility consists of enlarging the range of displacements of humans observed by a given acquisition setup. Indeed, when enlarging displacements relative to the center of the acquisition setup, some keypoints end up falling outside the viewpoints of some cameras. As the range of displacement in Human3.6M and CMU datasets is very limited, we run this experiment on the test set of AMASS dataset. It goes without saying that we do not use this test set in any of our training scenarios.
Here, we run MHP on both training and test set of AMASS four times, each of which considering different ranges of displacement for randomly positioning the human meshes in the scene. This translates to the room size parameter in MHP ranging from $0\times0$ to $3\times3$.
Then we train our MPL on each training set and test the trained model on the corresponding test set. We also run the baseline, \ie, triangulation, on the same test set. 
We consider a 2-view camera setup and run our test on all possible pairs of the 5 selected HD cameras (3, 6, 12, 13, 23) in CMU.
\cref{fig:amass-range-of-freedom} depicts how the range of displacements affects the keypoints' visibility (\cref{subfig:vis}) and the MPJPE (\cref{subfig:mpjpe}).
The figure confirms the robustness of MPL to missing keypoints compared to the baseline.

\end{document}